**MANUSCRIPT**

# Learning The Sequential Temporal Information with Recurrent Neural Networks

Pushparaja Murugan*

Correspondence:
pushparajam@xrvision.com
XRVision Research and Development center, Singapore,
*

**Abstract**

Recurrent Networks are one of the most powerful and promising artificial neural network algorithms to processing the sequential data such as natural languages, sound, time series data. Unlike traditional feed-forward network, Recurrent Network has a inherent feed back loop that allows to store the temporal context information and pass the state of information to the entire sequences of the events. This helps to achieve the state of art performance in many important tasks such as language modeling, stock market prediction, image captioning, speech recognition, machine translation and object tracking etc., However, training the fully connected RNN and managing the gradient flow are the complicated process. Many studies are carried out to address the mentioned limitation. This article is intent to provide the brief details about recurrent neurons, its variances and trips & tricks to train the fully recurrent neural network. This review work is carried out as a part of our IPO studio software module 'Multiple Object Tracking'.

## 1 Whats wrong with the Feed forward network

Almost every incident of our life is likely to be happening in sequential manner. If any of external incident disturb the state of the sequence, the compete sequence will be disturbed. For example, the human language, sequences of nodes in music, running, walking etc., In all those sequences of events, time defines the occurrence of events and the genome of sequences determine the event itself. In order to understand the readable outputs of the sequences, we need to have the prior knowledge about the sequential data of time series. If we want to create an artificial intelligence system to understand and predict the time series sequential events, traditional forward neural networks fails to address this issue. Though the feed forward network and convolutional neural networks perform well in classification problems, object detection, localization processing the sequential dynamic data cannot be done. The elaborated explanation of convolutional neural network is given in our previous articles [1], [2], [3], [4]. The neural unit called 'Recurrent Neural Network' address this issue. Wikipedia says that 'a recurrent neural network (RNN) is a class of artificial neural network where connections between units form a directed graph along a sequence. This allows it to exhibit dynamic temporal behavior for a time sequence'. Unlike feedforward neural networks, RNNs can use their internal state (memory) to process sequences of inputs. This makes them applicable to tasks such as unsegmented, connected handwriting recognition or speech recognition [5]. In other hand, the recurrent neural network has a typical internal loop connections that allow the information to persist and make use of sequential information. Consider a pattern recognition AI that uses the spatial information about the appeared objects in the image. In this simple classification framework, feed-forward neural network (typically CNN-Convolutional Neural Network) can be trained with multiple images with the concerned objects. The feed-forward neural network, known as multi-layer perceptron consists of an input layer, hidden layer and output layer. The input layer receives the information from the natural data, weighted sum of this information passes through the activation layers and output layer produce the output of added weighted sum of information added with bias values. Each hidden layers has own weights and bias. The non-linear activation layer break the linear relationship between the neural units. Hence all of the hidden layers behaves independently and the relationship between the successive layers will help us to understand the pattern. However, the independence between the neural units as in MLP led the network to incapable of persisting the information about the sequential events. The semantics of a typical Multi-layer preceptron is shown in the Figure 1.

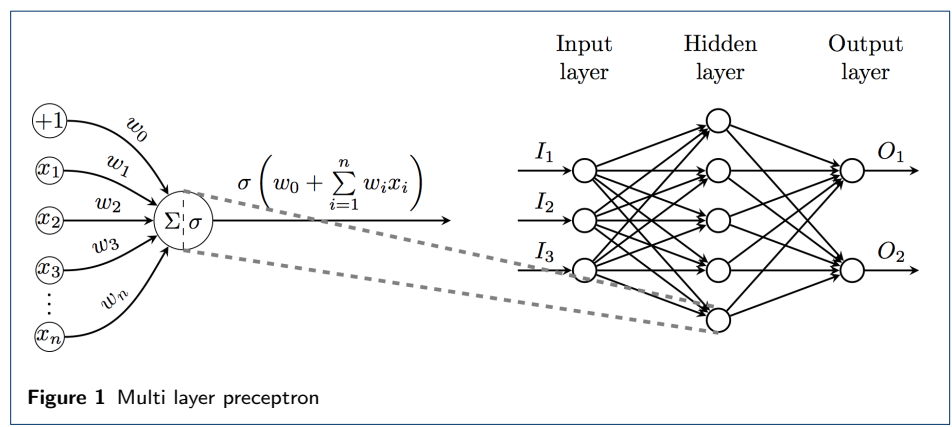

**Figure 1** Multi layer preceptron

## 2 Recurrent Neural Network - The rescuer

Dynamic networks are a variance of Feed-forward network that has the ability of processing temporal information to deal with dynamic operations. These dynamic neural networks, not only process the information but also the previous operations of the network. The processing of temporal information are classified into two groups, Feed-forward neural network with delay unit (Time-delay network) and Feed-forward neural network with feed-back loop (Recurrent Neural Network). The time delay network process some of inputs before the data is presented to neurons in the networks. The Recurrent neural networks that have recurrent connections as well as the structures of delay elements seen in time delay networks. Recurrent Neural Network can be non-linear or linear dynamic in nature. This distinction of these mentioned classes based on the representation of the time such as continuous and discrete in the present system and based on representation of the signal such as real-valued and quantized. However these properties do not completely implies about the application of the Recurrent Neural Networks. Based on the application, the Recurrent Neural Networks can be broadly can be used for

- Associative memories
- Sequence mapping systems

The main objective of the Recurrent Neural Networks is, to make use of the sequential information about the events. In other hand, the Recurrent Neural Networks generate the outputs based on the previous events by using closed loop connection with the unit cell itself to perceive the information about the past history that allows to process the temporal informations. So that the activation can flow through the closed loop. The figure 2 illustrates the example of Simple Recurrent Neural Network (SRN) proposed by Elman [6].

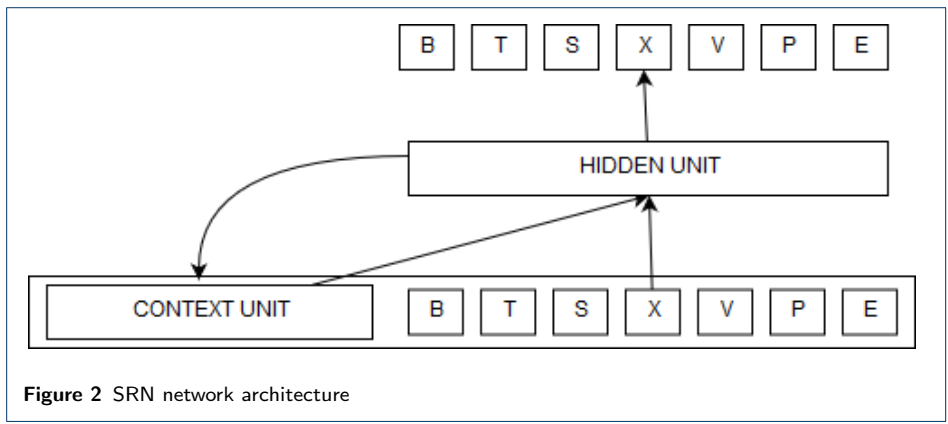

**Figure 2** SRN network architecture

Simple Recurrent Network has three layers, input units added with context unit, hidden units and output units. The feedback loop from the hidden units to the context units in the input units allow the network to process the previous stage of any sequential events. Hence, the network will have two sources of inputs such as input information and the state of previous events. Perceiving the stage of previous events is commonly known as the memory of Recurrent Neural Networks.

2.1 Vanilla Recurrent Neuron

Consider a single Recurrent Neuron as shown in the Figure 2.1. Temporal information, known as state about the successive events are passed through the non-linear activation function $\sigma$ and fed back to the unit cell itself. The recurrent network unit applies the recurrence formula to the input vector $i_t$ at time step $t$ and previous stage $s_{t-1}$ of the input vector at the time step $(t-1)$ and outputs the vector $x_{t+1}$ with new proposed stage $s_t$. The proposed stage can be expressed as,

$$s_t = \sigma(s_{t-1}, x_t; \theta) \tag{1}$$

where, $\theta$ is the learning parameter The hidden state $s_t \in \mathbb{R}^N$ at the time step $t$ is a function of input $x_t$ at the time step $(t-1)$. It can be computed by adding input information $x_t \in \mathbb{R}^M$ that is multiplied with weights $W_{iw} \in \mathbb{R}^{M \times N}$ at the time state $t$ with the previous state $(t-1)$ of input at the time step $(t-1)$ that is also multiplied with its own weights $W_{sw} \in \mathbb{R}^{N \times N}$. This can be expressed as,

$$s_t = \sigma(W_{iw} * x_t + W_{sw} * h_{t-1} + b) \tag{2}$$

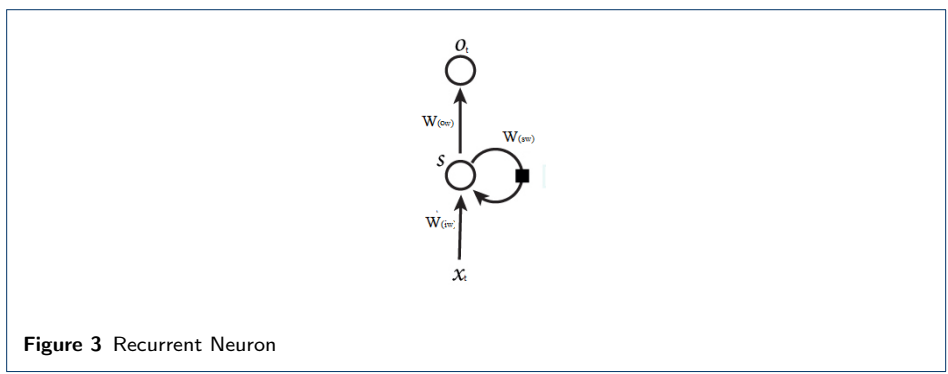

**Figure 3** Recurrent Neuron

The output of the recurrent neurons can be expressed as,

$$o_t = \sigma(W_{ow} s_t) \tag{3}$$

where $b \in \mathbb{R}^N$ is bias values.

## 2.2 Unfolding in time

The weighted sum of the inputs and the hidden states are transformed and squashed by the non-linear activation function. Non-linear transformation of the input will help the linear relationship between successive neural units and the squashing the values to small into a logistic space will help to works well for the gradients to flow smoothly on backpropagation.

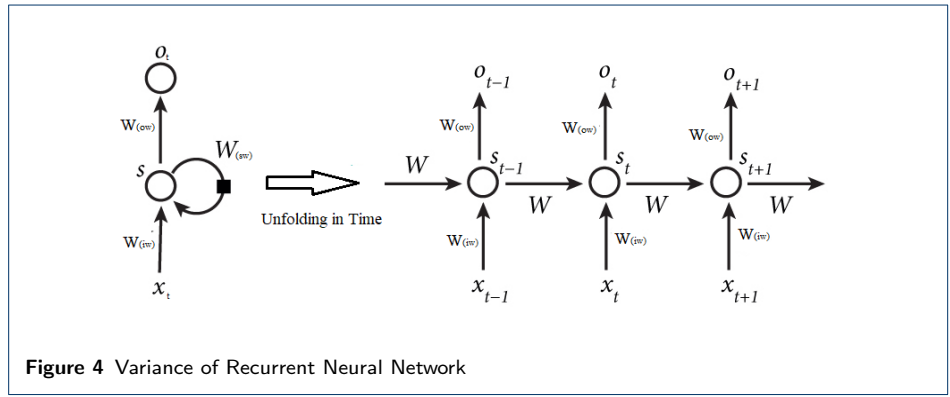

**Figure 4** Variance of Recurrent Neural Network

A typical recurrent neural networks is consisted of many recurrent unit in the same network and passes the time dependent information to the successive neural units. This implies that the recurrent neural network are highly related to sequences and lists. In over all, the hidden state as the memory of the networks captures the temporal information of the previous history that lead to computed outputs dependent on the network memory. Unlike the traditional feed-forward networks, the RNN shares same parameters across all the time steps that reduce larger number learning parameters. On the other hand, We are performing same task at each time step but with the different inputs.

## 3 Back Propagation Through Time

Lets consider that we are trying to minimize the distance between true value and predicted value by using cross entropy loss function. For a given true value $y_t$ at time step $t$, the predicted value is $\hat{y}_t$ the cost or loss function can be expressed as,

$$\mathbb{L}_t(y_t, \hat{y}_t) = -y_t \log(\hat{y}_t) \tag{4}$$

where $\mathbb{L}_t$ is the error values at the time $t$ for single values in the sequence. Hence the total error for recurrent nerual network unrolled in time is, sum of the errors at every time steps.

$$\mathbb{L}(y_t, \hat{y}_t) = -\sum_t (y_t, \hat{y}_t) \tag{5}$$

$$\mathbb{L}(y_t, \hat{y}_t) = -\sum_t y_t \log(\hat{y}_t) \tag{6}$$

The learning of the network can be done by minimizing the negative log likelihood. That can be done by adjusting the parameters of the networks with back-propagating the error through the network. Back propagation is computing gradients of expressions through recursive application of chain rule. Hence the learning of the network of Recurrent Neural Networks can be explained as computing the gradient of error with respect the parameters $(u, v, w, b; \theta)$. Unlike the feed-forward network, the weights are shared between all the sequences, hence we can differentiate and sum together. Hence the total gradient with respect to the weight $w_{iw}$ at the time step can be written as,

$$\frac{\partial \mathbb{L}}{\partial W_{iw}} = \sum_t \frac{\partial \mathbb{L}_t}{\partial W_{iw}} \tag{7}$$

Chain rule will help us to update the learning parameters in the network, ie,

$$\frac{\partial \mathbb{L}}{\partial W_{iw}} = \sum_t \frac{\partial \mathbb{L}_t}{\partial z_t} \frac{\partial z_t}{\partial w_{iw}} \tag{8}$$

where $z_t$ is the hidden layers weighted sum of the information passed through that layer. Derivative of bias can be expressed as,

$$\frac{\partial \mathbb{L}}{\partial b_{iw}} = \sum_t \frac{\partial \mathbb{L}_t}{\partial z_t} \frac{\partial z_t}{\partial b_{iw}} \tag{9}$$

Lets consider the back propagation through the temporal information transferring connection. For a given time step $t$, the partial derivatives of the loss function with respect to the weights of state information connection $w_{sw}$ can be written as,

$$\frac{\partial \mathbb{L}_t}{\partial w_{sw}} = \frac{\partial \mathbb{L}_t}{\partial z_t} \frac{\partial z_t}{\partial s_t} \frac{\partial s_t}{\partial w_{sw}} \tag{10}$$

From this, We can yield the back propagation with respect to the weights $w_{sw}$ from the time step $t$ to 0 can be expressed as,

$$\frac{\partial \mathbb{L}_t}{\partial w_{sw}} = \sum_1^t \frac{\partial \mathbb{L}_t}{\partial z_t} \frac{\partial z_t}{\partial s_t} \frac{\partial s_t}{\partial w_{sw}} \tag{11}$$

Hence the gradient of the loss function with respect to the weights $w_{sw}$ for the whole sequences is,

$$\frac{\partial \mathbb{L}_t}{\partial w_{sw}} = \sum_{t-1} \sum_1^t \frac{\partial \mathbb{L}_t}{\partial z_t} \frac{\partial z_t}{\partial s_t} \frac{\partial s_t}{\partial s_{t-1}} \frac{\partial s_{t-1}}{\partial w_{sw}} \tag{12}$$

From the Eq. (12), we can clearly understand that the state of information about the sequences is preserved in the current state. RNNs have a sense some memory about past history of sequence of data. This helps the system to gain context information. Theoretically RNNs have infinite memory, that implies that the capabilities of look back indefinitely. However, practically they can only look back a last few steps.

### 3.1 Forward propagation and gradient flow in variances of RNN

Depending on the field of application, recurrent neural network can be unfold in variety of ways. Common architecture are One to one, one to many, many to one and many to many. Each network architecture serves state of art performance in many complex tasks. Different types of fully connected RNN is shown in the Figure 3.1. The weight updates and the gradient flow of these architecture are shown in the Figures 3.1 3.1 3.1 3.1.

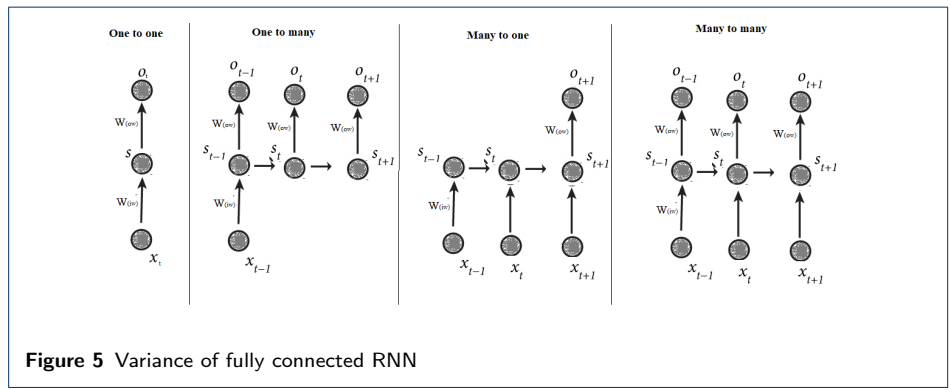

**Figure 5** Variance of fully connected RNN

## 4 Fall of Vanilla Recurrent Neurons

Consider the back propagation over the temporal information connections in Eq (12), the parameters are shared across the network. It actually helps RNN to learn long term dependencies. However, the partial derivative $\frac{\partial s_t}{\partial s_{t-1}}$ does also follow the chain rule to compute such that $\frac{\partial s_t}{\partial s_{t-1}} = \frac{\partial s_t}{\partial s_{t-2}} \frac{\partial s_{t-2}}{\partial s_{t-1}}$. Since the network requires to back propagate the gradient of loss function $\frac{\partial \mathbb{L}}{\partial \theta}$ for long sequences, gradient with respect to each time step has to be computed and need to be added. So the value of the gradient itself shrink in values and tends to vanish after few steps. Hence the state far away from the current steps does not contribute to the parameters' gradient computing. Another issue is the gradient exploding, which attributed to large values in matrix multiplication. It shows that, the RNN has series difficulties in learning long-range dependencies. Though the state of information in any real event sequence changes cross the time steps and have large influence in the present state, this can be extremely insensitive in RNN due to difficulties in learning long term dependencies.

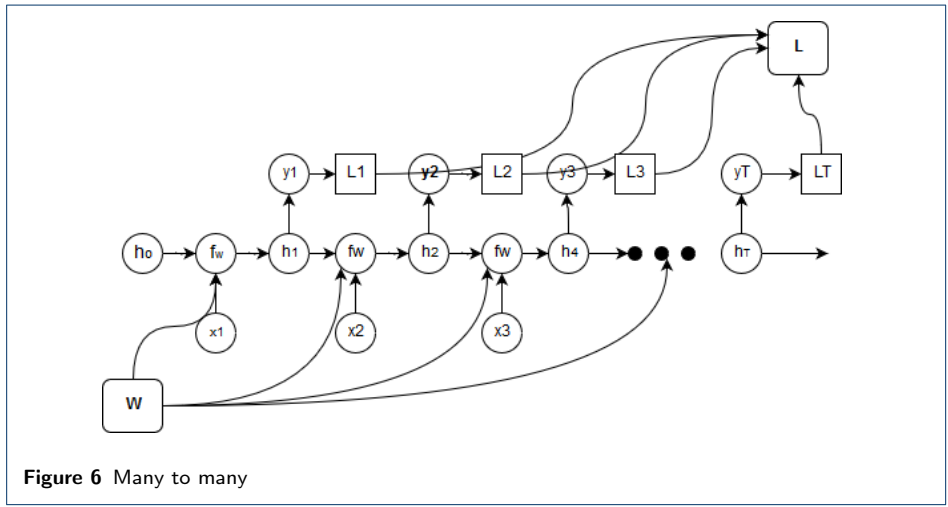

**Figure 6** Many to many

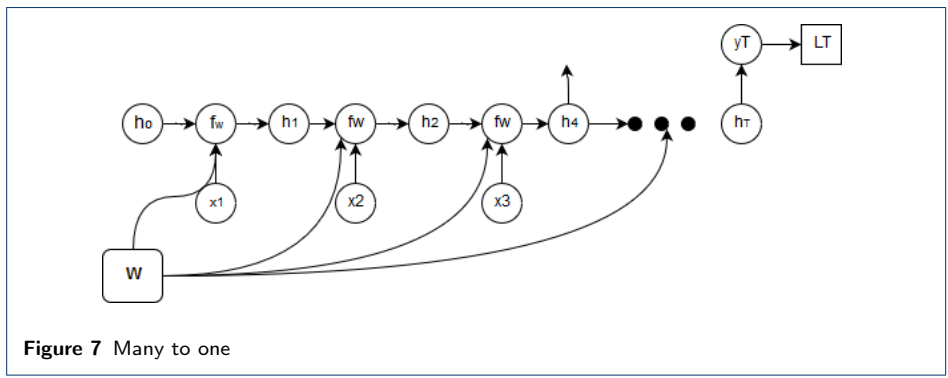

**Figure 7** Many to one

Vanishing gradient problem affects the RNN effecitiveness is limited when it needs to go back deeper in the context. There is no finer control over which part of the context needs to be carried and how much of the past needs to be forgotten.

## 5 Combat with vanishing and exploding gradients

Practically, Training the fully connected RNN and managing the gradient flow hardest part. The following trips and tricks helps to manage the gradient flow [7] [8] [9].

5.1 Architecture and Memory core
- LSTM and GRU core cell are profoundly promising alternative for RNN hidden units to combat with vanishing and exploding gradients.
- Addition of Layer normalization to the linear mapping for the RNN helps to speed up the learning as well as increase the performance.

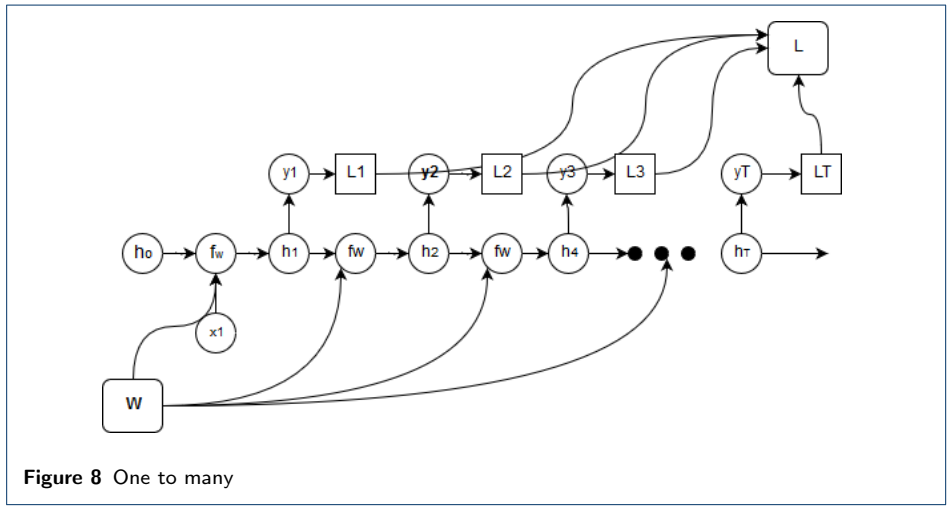

**Figure 8** One to many

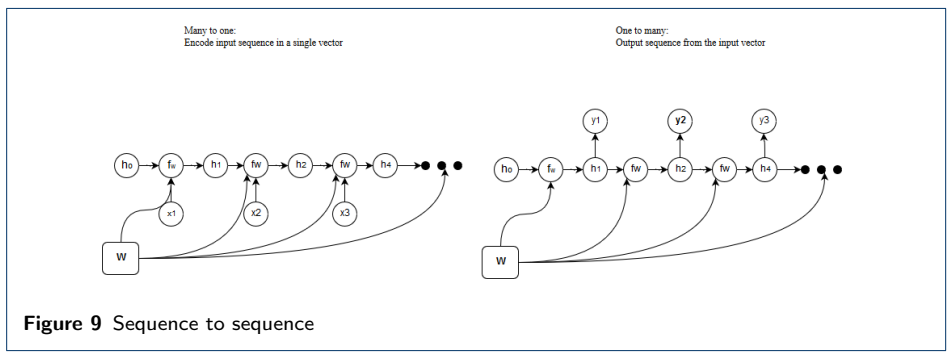

**Figure 9** Sequence to sequence

- Preprocessing of the input with feed forward layer helps to project the data into the space with temporal dynamics which leads to improve the performance.
- Stacking the RNN network in the out of plane and summing up the outputs help to increase the performance since the RNN requires quadratic number of weights.

5.2 Architecture parameters

- Careful initialization of learning parameters such as training the initial state as the variable can largely improve the performance.
- Initialization of forget gate bias to 1 by default helps to improve the performance as the RNN learn time to learn the usual information from the past.
- Implementation of regularization technique such as Dropout

5.3 Learning parameters

- Using Adaptive learning rate such as 'RMSprop' and 'ADAM' optimizer depends on the gradient value to handle the complex training dynamics
- Gradient clipping to scale down the gradient to avoid the gradient exploding
- Normalizing the loss to obtain the loss with similar magnitude in all over the datasets. It can be done by summing the loss values along the sequences and dividing by max length of sequence.

- Truncated back-propagation. Instead of back propagation the error value from the time step $t$ to time step 0, training the squences $n$ number of steps help to preserve the state of information
- Training the network for long period of time.

## 6 Rise of LSTM neuron

LSTM Neuron is just an another variance of Recurrent Neuron, also known as Memory block of LSTM network. It is the fundamental basic unit of fully connected LSTM network. It was introduced by Hochreiter & Schmidhuber (1997) to overcome the limitation of vanilla RNN explicitly by providing the separate pipe line for gradient flow for longer time steps. The LSTM neurons has separate Cell state to preserve the long term temporal information of the sequences. Each neuron has pair of adaptive and multiplicative gates which input and output to the neuron. Also each neurons has a core and recurrently self connected linear unit called as 'constant error carousel' and its activation is known as the 'Cell state'.

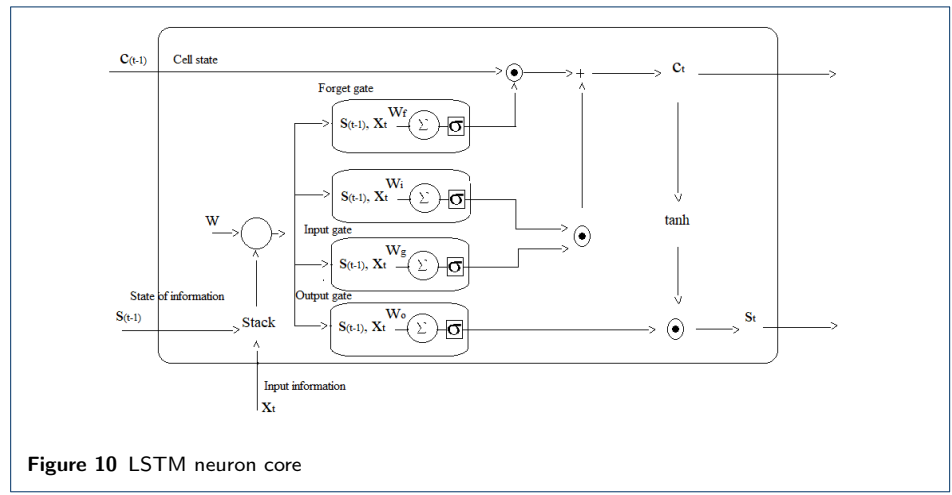

**Figure 10** LSTM neuron core

On other hand, Unlike vanilla RNN, LSTM has four neural units inside and interact with each other in special way where the RNN has only one neural unit and the output values are squished between -1 to 1 by $tanh$ activation function. Practically, the architecture of LSTM cell is compared with the conveyor belt which allows the information to flow without any changes but with small manipulation. The typical hierarchy structure of the LSTM allows the information to the cell state if it needed and flush out when it was not needed. The gates are nothing but a composed of a sigmoid acitvation and a tiny affine layers which allows the information to pass optionally. sigmoid activation non-linearity used to squash the output values between zero and one in order to decide how much of information of each component should be through. The output zero and one describes the state information is passed complete to the cell state and flushed out from the gate itself. A typical LSTM has three gates to protect and govern the cell state. They are

- Forgot gate
- Input gate
- Output gate

### 6.1 Forgot gate

The forget gate in LSTM architecture is responsible for flushing out the information that no longer required and adding the important information to flow in the Cell state. The forget gate decide how much information to pass to the cell state through the sigmoid activation function. Squashing the value between zero to one implies that how much of previous state should be forgotten and also Manipulates how much of previous cell state should be kept within the update to next state. The input information $x_t$ at time step $t$ and the previous state of information $h_{t-1}$ is multiplied with weight matrix $w_{if}$ and $w_{sf}$ and added with bias $b_f$. This value of the affine layer is passed and directly multiplied with the cell state $s_t$ at the time step $t$. This can be expressed as follow,

$$O_f = \sigma[h_{t-1} * w_{sf} + x_t * w_{if} + b_f] * s_t \qquad (13)$$

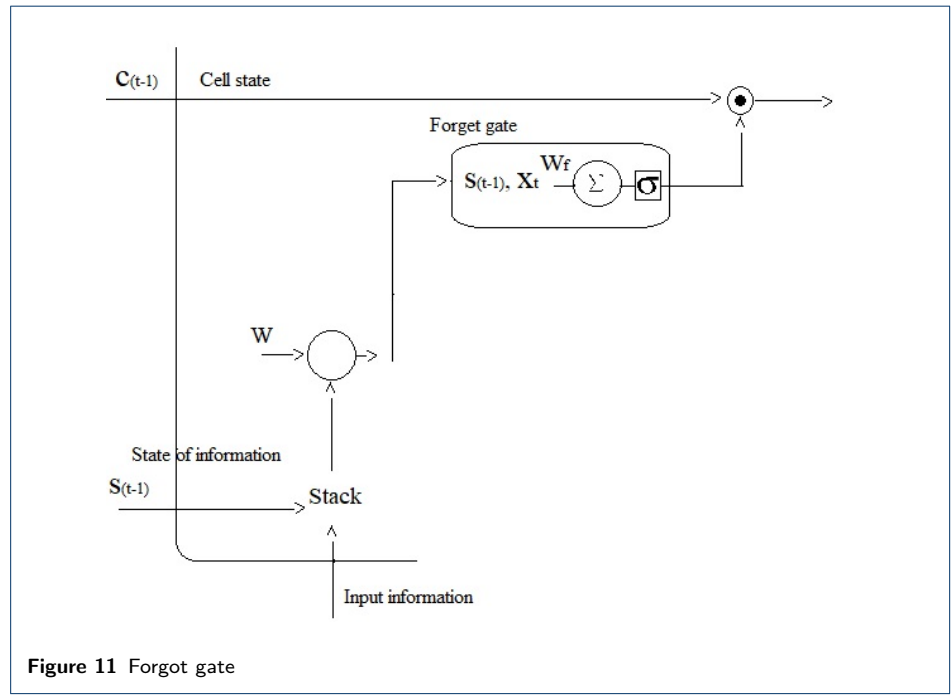

**Figure 11** Forgot gate

The forget gate depends on the previous state of information $ht-1$ and input $x_t$, and outputs a value squashed between 0 and 1 for each number in the cell state $c_t$. 1 represents completely keep this while a 0 represents "completely get rid of this. As soon as the forget layer encounters the change of context in the input information, the subject of the value is completely flushed out and new subject is added to the cell state. This is highly required for optimizing the performance of the LSTM network overall.

### 6.2 Input gate

The objective of the input gate is to decide what new information that needs to be stored in the cell state. The function of the input gate is two parts namely and input gate layer and a information proposal layer. The input gate layer is the multiplied values of weight matrix and previous state information and current input information and then the values are squished by sigmoid activation. Next the proposal layers proposes new input candidate value to the cell state by squashing the multiplied and added values of weight matrix, previous state information and current input information. These combined value create the next update for the cell state.

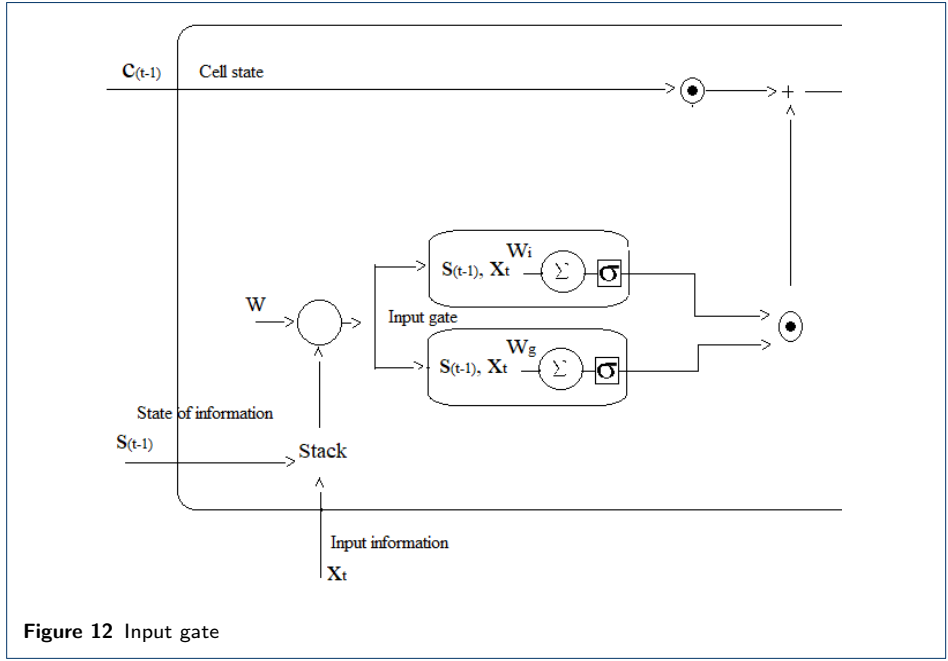

**Figure 12** Input gate

The update of the cell state $s_{t+1}$ from $s_t$ can be expressed as follows, If the previous state $h_{t-1}$ and the current input $x_t$ is passed through sigmoid acitvation function after multipied and added weight and bias respectuvely, the output can be given as,

$$O_{il} = \sigma[h_{t-1} * w_{sil} + x_t * w_{iil} + b_{il}] \qquad (14)$$

The new proposal of input candidate can be expressed as,

$$O_{if} = \sigma[h_{t-1} * w_{sif} + x_t * w_{iif} + b_{if}] \qquad (15)$$

Hence the update of the next cell state $s_{t+1}$ can be expressed as,

$$s_{t+1} = O_{il} + O_{if} \qquad (16)$$

In other hand, the update of old cell state $s_t$ into the new cell state $s_{t+1}$ can be explained by following process.
- Regulating the input information $x_t$ at the time step that has to be updated in next cell state $s_{t+1}$ at the time step $t+1$.
- Proposing new input information $o_{if}$ based on the previous state input information $h_{t-1}$ and current input information $x_t$.

- Updating the cell state $s_{t+1}$ at the time step $t+1$ by adding the regulated and proposed new inputs together.

### 6.3 Output gate

Based on the cell state, the output gate decides what is the most desired outputs to pass. The output of the output gate is obtained by the cell state of the LSTM passed through $tanh$ activation to squash the value between -1 to 1 and multiplied by the output gate of the sigmoid non-linearity. The function of the output gate is shown in the Figure 6.3. The state $s_t$ of the LSTM cell at the time step $t$ for a given cell state $c_t$ can be obtained by multiplying the output of $tanh$ and the output of sigmoid of the affine layer. The affine layer depends on the previous state of the LSTM $s_{t-1}$ and the input information. This can be expressed as follows, The output of the affine layer $o_t$ at the time step $t$ is multiplying the input $x_t$ at the time step $t$ and previous state of LSTM $h_{t-1}$ at the time step $t-1$ with the weight matrix $w_{io}$ and $w_{so}$ respectively.

$$O_t = \sigma[h_{t-1} * w_{so} + x_t * w_{io} + b] \tag{17}$$

where $b$ is bias values Hence the final output of the output layer can be expressed, as the output of affine layer multiplied with $tanh$ output of the cell state,

$$h_t = O_t * tanh(s_t) \tag{18}$$

The main objective of outputting the useful information can be broken down into three steps.
- Scaling down the value of the cell state between -1 and +1
- Adding the value of sigmoid function squished affine layer's value to regulate the cell state information
- The above step ensure that it diminishes all other values except the important. Thus it needs to be built on the input and hidden state values and be applied on the cell state vector to get the gain scaling properties.

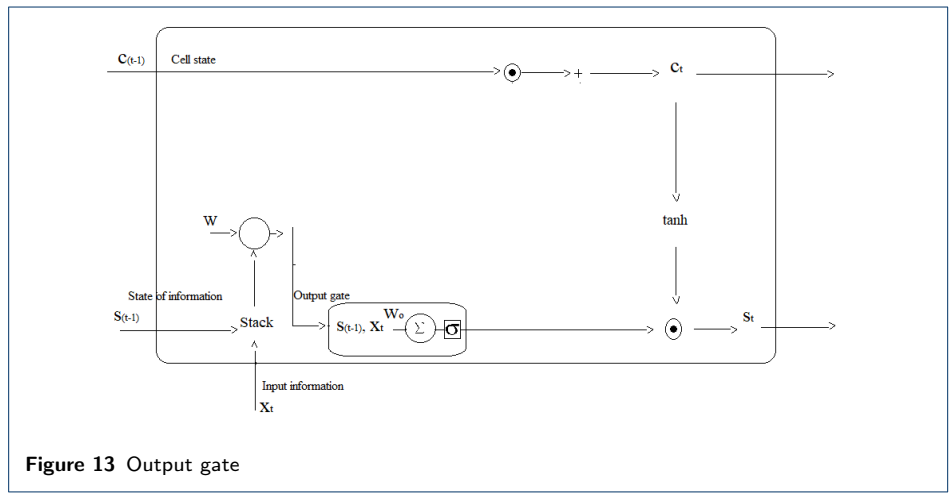

**Figure 13** Output gate

## 7 Simplified LSTM - The GRUs

Another important RNN variance to overcome the limitation of vanishing and exploding gradient is GRU, known as Gate Recurrent Unit. It is an improved version of LSTM. However, Unlike LSTM cell, GRU has only two gates to preserve the temporal information of the sequential data namely Update gate and reset gate. Another important thing about GRU, it can be trained for very long sequences without flushing out and removing the irrelevant information of the sequences with minimum implementation of learning parameters as compares with LSTM. It combines the forget and input gate to single 'update gate and merges cell state and hidden state. Result of which, the GRU architecture is considerably simpler than LSTM architecture. The GRU cell takes input as previous state $s_{t-1}$ and current input $x_t$ and outputs the next update for the time step $t+1$. Just like the LSTM cell, the reset gate and update gate in GRU, the previous state information and current inputs are multiplied with weight matrix and added with bias values. It proposes new input candidate and state update for the next GRU cell.

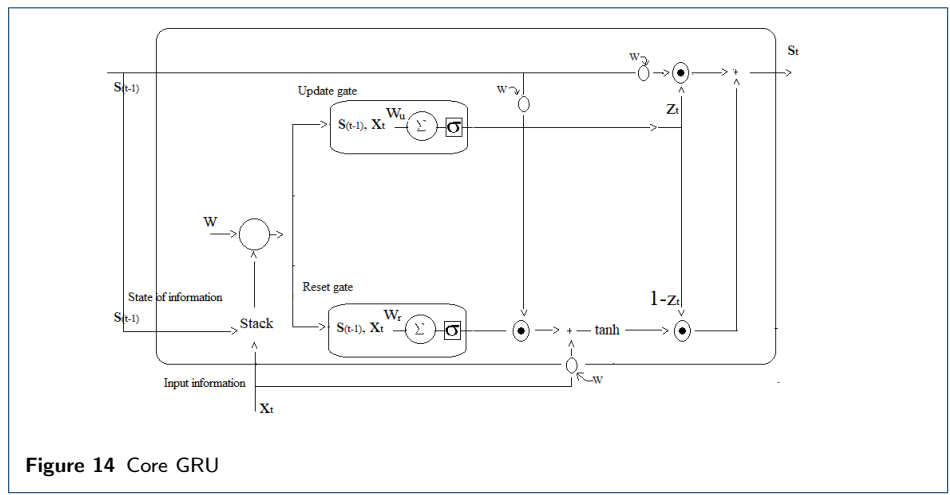

**Figure 14** Core GRU

### 7.1 Update gate

The update uses the current input information $x_t$ and previous state of information $s_{t-1}$ and own weight matrix to generate next proposed output. Current input and previous state information are multiplied by own weight matrix, added together and passed through sigmoid activation function. That helps to compute how much past information to pass to the future update.

$$z_t = \sigma[h_{t-1} * w_{zs} + x_t * w_{zi} + b] \tag{19}$$

### 7.2 Reset Gate or read gate

The reset gate decides how much of information previous state of information to forget and pass to the update. The reset gate takes the previous state of information $s_{t+1}$ and current input information $x_t$ as input and decides the value of forgetting based on the sigmoid activation function.

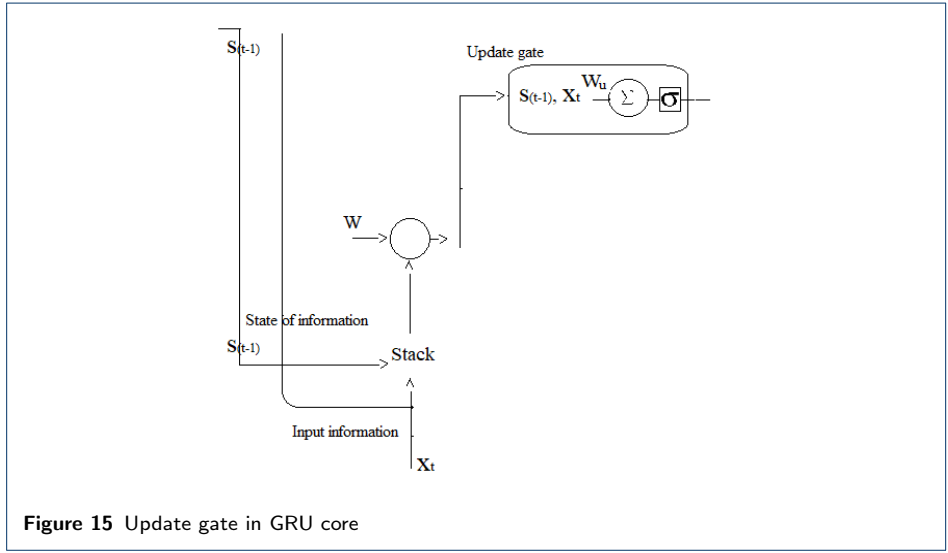

**Figure 15** Update gate in GRU core

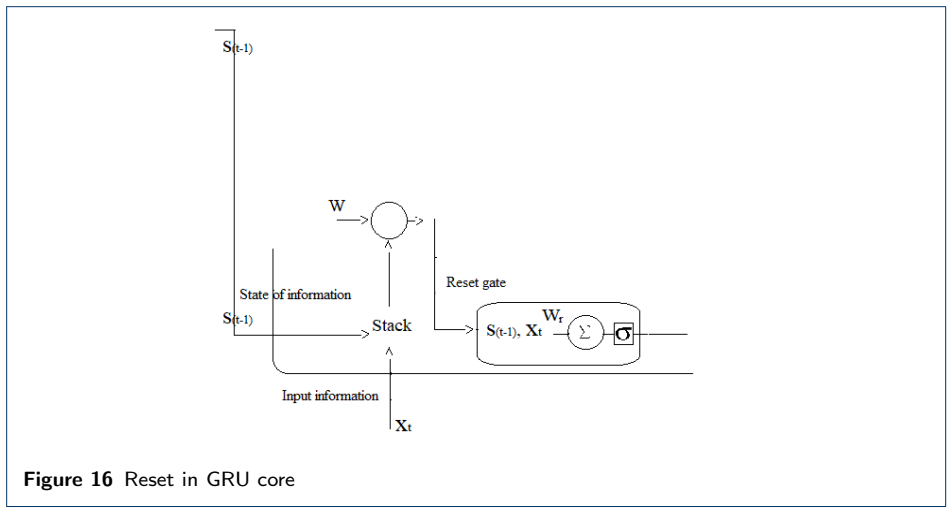

**Figure 16** Reset in GRU core

$$r_t = \sigma[h_{t-1} * w_{rs} + x_t * w_{ri} + b] \quad (20)$$

7.3 Current memory content

The update gate and reset gate manipulates the state of information and passes to future updates. The relevent information from the past can be decided by the reset gate by introducing new affine layer. On other hand, weight current input and weighted previous state of information can be obtained by multiplying with own weight matrix.The Hadmard product between the weighted input information & the reset gate output $r_t$ added with weighted previous state of information and passed through the $tanh$ activation function to get the current memory content.

$$h'_t = \tanh[x_t * w_{ci} + r_t \circledast h_{t-1} * w_{si} + b] \tag{21}$$

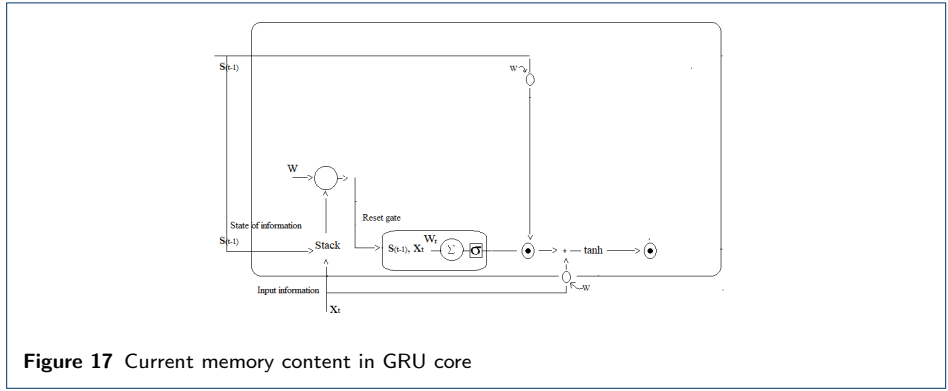

**Figure 17** Current memory content in GRU core

### 7.4 Future update

The future update or the final memory at the current time steps can be computed by adding the required value current memory context from the update gate and the required value from previous state of input information. The required value of current memory content is computed by taking Hadmard product between output of the update gate $z_t$ and previous state of input information $s_{t-1}$ and required value of previous state of input information is computed by Hadmard product between $1 - z_t$ and the current state of input information $s_t$. Added value of these two gives the future update for the next GRU core.

$$h_t = [z_t \circledast s_{t-1} + (1 - z_t) \circledast s_{t-1}] \tag{22}$$

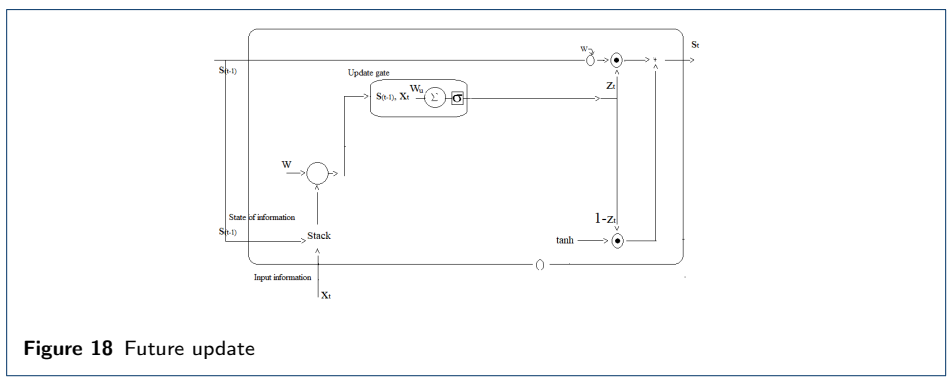

**Figure 18** Future update

## 8 Conclusion

Though numerous variance of Recurrent neurons are invented, this article is only intended to discuss about the popular architecture. The role of recurrent neuron in processing sequential data is discussed briefly and also the limitation of recurrent neurons is addressed. The core modules of LSTM cell and GRU cells are elaborately explained.

**Acknowledgement**

This review work is carried out as a part of our IOP studio software development / Multiple Object Tracking Module. We thank our CEO Mr. Guy Ron and our team lead Mr. Shree Ramakrishnan for the support.